\documentclass[tablecaption=bottom,wcp,algo2e,ruled,plain,noend,vlined,linesnumbered]{jmlr}

\PassOptionsToPackage{numbers,sort&compress,sectionbib}{natbib}
\usepackage{chapterbib}

\usepackage{filehook}

\usepackage[utf8]{inputenc}         
\usepackage[T1]{fontenc}            
\usepackage{url}                  	
\usepackage{doi}           	 		
\usepackage{booktabs}       		
\usepackage{amsfonts}       		
\usepackage{nicefrac}       		
\usepackage{microtype}      		
\usepackage{lipsum}                 
\usepackage{titlesec}
\usepackage{booktabs}
\usepackage{enumitem}
\usepackage{float}

\usepackage{amssymb,amsmath}
\usepackage{mathtools}
\usepackage{chngcntr}

\usepackage{dsfont}

\usepackage{setspace}
\usepackage{listings}
\usepackage{fancyvrb}
\fvset{fontsize=\small}

\newcommand{\myeqp}[1]{\hyperref[eq:#1]{Eq.\ref*{eq:#1}}}
\newcommand{\mysec}[1]{\hyperref[sec:#1]{Section~\ref*{sec:#1}}}
\newcommand{\mytable}[1]{\hyperref[table:#1]{Table~\ref*{table:#1}}}
\newcommand{\myfig}[1]{\hyperref[fig:#1]{Fig.~\ref*{fig:#1}}}
\newcommand{\myappendix}[1]{\hyperref[appendix:#1]{Appendix~\ref*{appendix:#1}}}
\newcommand{\myalg}[1]{\hyperref[alg:#1]{Algorithm~\ref*{alg:#1}}}
\newcommand{\mytheorem}[1]{\hyperref[theorem:#1]{Theorem~\ref*{theorem:#1}}}
\newcommand{\myfootnote}[1]{\hyperref[footnote:#1]{Footnote~\ref*{footnote:#1}}}

\newcommand{\mathbold}[1]{\ensuremath{\boldsymbol{\mathbf{#1}}}}

\DeclareMathOperator*{\argmax}{arg\,max}

\newcommand{\bx}{\mathbold{x}}

\newcommand{\bphi}{\mathbold{\phi}}

\newcommand{\btheta}{\mathbold{\theta}}

\newcommand{\bSigma}{\mathbold{\Sigma}}

\newcommand{\bbE}{\mathbb{E}}
\newcommand{\bbH}{\mathbb{H}}
\newcommand{\bbI}{\mathbb{I}}

\newcommand{\bbV}{\mathbb{V}}

\newcommand{\mcD}{\mathcal{D}}
\newcommand{\mcL}{\mathcal{L}}

\newcommand{\rmd}{\mathrm{d}}

\newcommand{\GP}{\mathcal{G}\mathcal{P}}

\newcommand{\param}{\mathbold{\theta}}
\newcommand{\data}{\mathbold{x}}
\newcommand{\obsdata}{\data_o}

\newcommand{\hlik}{\hat{\mcL}}

\newcommand{\hp}{\hat{p}}
\newcommand{\tp}{\tilde{p}}

\DeclareUnicodeCharacter{00C4}{\"A}
\DeclareUnicodeCharacter{00E4}{\"a}

\DeclareUnicodeCharacter{00A0}{~}
\DeclareUnicodeCharacter{2010}{-}
\DeclareUnicodeCharacter{2013}{--}
\DeclareUnicodeCharacter{2014}{---}
\DeclareUnicodeCharacter{2009}{\,}
\DeclareUnicodeCharacter{2026}{\textellipsis}

\DeclareUnicodeCharacter{0391}{A}
\DeclareUnicodeCharacter{0392}{B}
\DeclareUnicodeCharacter{0393}{\Gamma}
\DeclareUnicodeCharacter{0394}{\Delta}
\DeclareUnicodeCharacter{0395}{E}
\DeclareUnicodeCharacter{0396}{Z}
\DeclareUnicodeCharacter{0397}{H}
\DeclareUnicodeCharacter{0398}{\Theta}
\DeclareUnicodeCharacter{0399}{I}
\DeclareUnicodeCharacter{039A}{K}
\DeclareUnicodeCharacter{039B}{\Lambda}
\DeclareUnicodeCharacter{039C}{M}
\DeclareUnicodeCharacter{039D}{N}
\DeclareUnicodeCharacter{039E}{\Xi}
\DeclareUnicodeCharacter{039F}{O}
\DeclareUnicodeCharacter{03A0}{\Pi}
\DeclareUnicodeCharacter{03A1}{P}
\DeclareUnicodeCharacter{03A3}{\Sigma}
\DeclareUnicodeCharacter{03A4}{T}
\DeclareUnicodeCharacter{03A5}{Y}
\DeclareUnicodeCharacter{03A6}{\Phi}
\DeclareUnicodeCharacter{03A7}{X}
\DeclareUnicodeCharacter{03A8}{\Psi}
\DeclareUnicodeCharacter{03A9}{\Omega}

\DeclareUnicodeCharacter{03B1}{\alpha}
\DeclareUnicodeCharacter{03B2}{\beta}
\DeclareUnicodeCharacter{03B3}{\gamma}
\DeclareUnicodeCharacter{03B4}{\delta}
\DeclareUnicodeCharacter{03B5}{\epsilon}
\DeclareUnicodeCharacter{03B6}{\zeta}
\DeclareUnicodeCharacter{03B7}{\eta}
\DeclareUnicodeCharacter{03B8}{\theta}
\DeclareUnicodeCharacter{03B9}{\iota}
\DeclareUnicodeCharacter{03BA}{\kappa}
\DeclareUnicodeCharacter{03BB}{\lambda}
\DeclareUnicodeCharacter{03BC}{\mu}
\DeclareUnicodeCharacter{03BD}{\nu}
\DeclareUnicodeCharacter{03BE}{\xi}
\DeclareUnicodeCharacter{03BF}{\omicron}
\DeclareUnicodeCharacter{03C0}{\pi}
\DeclareUnicodeCharacter{03C1}{\rho}
\DeclareUnicodeCharacter{03C3}{\sigma}
\DeclareUnicodeCharacter{03C4}{\tau}
\DeclareUnicodeCharacter{03C5}{\upsilon}
\DeclareUnicodeCharacter{03C6}{\phi}
\DeclareUnicodeCharacter{03C7}{\chi}
\DeclareUnicodeCharacter{03C8}{\psi}
\DeclareUnicodeCharacter{03C9}{\omega}

\DeclareUnicodeCharacter{0229}{\c{e}}

\jmlrvolume{96}
\jmlryear{2018.}
\jmlrsubmitted{02/15/2019}
\jmlrpublished{publication date}
\jmlrworkshop{1st Symposium on Advances in Approximate Bayesian Inference}
\jmlrproceedings{AABI 2018}{Proceedings of Machine Learning Research}

\title{Likelihood-free inference with emulator networks}

\author{\Name{Jan-Matthis Lueckmann}$^{1,2}$ \Email{jan-matthis.lueckmann@caesar.de} \\
        \Name{Giacomo Bassetto}$^{1,2}$ \Email{giacomo.bassetto@caesar.de} \\
        \Name{Theofanis Karaletsos}$^3$ \Email{theofanis@uber.com} \\
        \Name{Jakob H. Macke}$^{1,2,4}$ \Email{macke@tum.de}}

\begin{document}

\maketitle

\begin{abstract}
Approximate Bayesian Computation (ABC) provides methods for Bayesian inference in simulation-based models which do not permit tractable likelihoods. We present a new ABC method which uses probabilistic neural \emph{emulator} networks to learn synthetic likelihoods on simulated data -- both `local' emulators which approximate the likelihood for specific observed data, as well as `global' ones which are applicable to a range of data. 
Simulations are chosen adaptively using an acquisition function which takes into account uncertainty about either the posterior distribution of interest, or the parameters of the emulator. 
Our approach does not rely on user-defined rejection thresholds or distance functions. We illustrate inference with emulator networks on synthetic examples and on a biophysical neuron model, and show that emulators allow accurate and efficient inference even on problems which are challenging for conventional ABC approaches.
\end{abstract}

\stepcounter{footnote}\footnotetext{Computational Neuroengineering, Department of Electrical and Computer Engineering, Technical University of Munich, Germany}
\stepcounter{footnote}\footnotetext{Neural Systems Analysis, Research Center caesar, an associate of the Max Planck Society, Bonn, Germany}
\stepcounter{footnote}\footnotetext{Uber AI Labs, Uber Technologies, Inc., San Francisco, CA}
\stepcounter{footnote}\footnotetext{Part of this work was done while J.H.M was at the Centre for Cognitive Science, Technische Universität Darmstadt, Germany}

\setcounter{footnote}{0}

\section{Introduction}
\label{introduction}

Many areas of science and engineering make extensive use of complex, stochastic, numerical simulations to describe the structure and dynamics of the processes being investigated \cite[]{Karabatsos_2017}. A key challenge in simulation-based science is linking simulation models to empirical data: Bayesian inference provides a general and powerful framework for identifying the set of parameters which are consistent both with empirical data and prior knowledge. One of the key quantities required for statistical inference, the likelihood of observed data given parameters, $\mathcal{L}(\param)=p(\obsdata|\param)$, is typically intractable for simulation-based models, rendering conventional statistical approaches inapplicable. 

Approximate Bayesian Computation (ABC) aims to close this gap \cite[]{BeaumontZhang2002}, but classical algorithms \cite[]{Pritchard1999, Marjoram2003} scale poorly to high-dimensional non-Gaussian data, and require \emph{ad-hoc} choices (i.e., rejection thresholds, distance functions and summary statistics) which can significantly affect both computational efficiency and accuracy. In \emph{synthetic likelihood} approaches to ABC \cite[]{Wood2010,OngNott2016,Price2018}, one instead uses density estimation to approximate the likelihood $p(s(\obsdata)|\param)$ on summary statistics $s(\cdot)$ of simulated data. A recent proposal by \cite{Jarvenpaa2017}, \cite{Gutmann2018} uses a Gaussian process ($\GP$) to approximate the distribution of the discrepancy $d(s(\data), s(\obsdata))$ as a function of $\param$, and Bayesian Optimization to propose new parameters. While this approach can be very effective even with a small number of simulations, it still requires summary statistics, choice of a distance function $d(\cdot, \cdot)$, and relies on assuming a homoscedastic $\GP$. 

The goal of this paper is to scale synthetic-likelihood methods to multivariate and (potentially) non-Gaussian, heteroscedastic data. We use neural-network based conditional density estimators \cite[which we call `emulator networks', inspired by classical work on emulation methods;][]{Kennedy_OHagan_2001}, to develop likelihood-free inference algorithms which are efficient, flexible, and scale to high-dimensional observations. Our approach does not require the user to specify rejection thresholds or distance functions, or to restrict oneself to a small number of summary statistics.

\begin{figure}
\label{fig:overview}
    \includegraphics[trim={0 0.15cm 0 0},clip,width=1.0\textwidth]{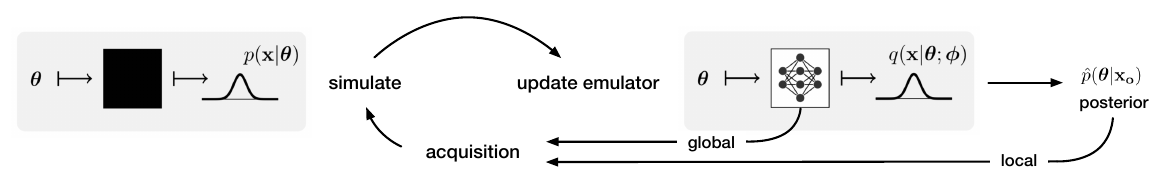}
    \caption{
    {\bf Likelihood-free inference with emulator networks.}
    Our goal is to perform approximate Bayesian inference on simulator-models, i.e. models from which we can generate samples, but for which we can not evaluate likelihoods.  We learn a tractable probabilistic emulator  $q(\data|\param; \phi)$ approximating the simulator $p(\data|\param)$. The emulator then serves as a synthetic likelihood to obtain an approximate posterior. To train the emulator using a low number of simulations, we use active learning to select informative samples: The acquisition rule is either based on the current posterior estimate (if observed data $\obsdata$ is given, `local' learning), or on our uncertainty about the  weights of the emulator network (`global' learning).}
\end{figure}

\section{Likelihood-free inference with emulator networks}

Our goal is to obtain an approximation to the true posterior $p(\param|\obsdata)$ of a black-box simulator model, i.e. models from which we can generate samples $\data \sim p(\data|\param)$, but for which we cannot evaluate likelihoods $\mcL(\param)$. To solve this task, we learn a synthetic likelihood function $\hlik(\param)$ by training a conditional density estimator on simulated data. We actively propose parameters for simulations, since simulations are often the dominant cost in ABC: Therefore, we want to keep the number of calls to the simulator as low as possible (\myfig{overview}).

Core to our approach is an emulator $q(\data|\param; \bphi)$, a conditional density estimator with parameters $\bphi$ that approximates the simulator $p(\data|\param)$. Having collected an initial simulated dataset $\mcD$, e.g. by repeatedly drawing from the prior $p(\param)$ and simulating data, the emulator is trained. 
We actively select new locations $\param^*$ for which to simulate new data points $\mcD^{*}=\{(\param^*, \data^*)\}$ to keep the number of calls to the (potentially computationally expensive) simulator low. $\mcD^{*}$ is appended to the dataset, the emulator is updated, and the active learning loop repeats. The emulator defines a synthetic likelihood function $\hlik(\param)=q(\data=\obsdata|\param; \bphi)$ that we use to find an approximate posterior, which is proportional to $\tp(\param|\obsdata):=\hlik(\param) p(\param)$. This approach is summarized in \myappendix{algorithm} in form of an algorithm.




Thus, our approach requires (1) an emulator, i.e., a flexible conditional density estimator, (2) an approach for learning the emulator on simulated data and expressing our uncertainty about its parameters, (3) an acquisition rule for proposing new sampling locations, and (4) an inference procedure for obtaining the posterior distribution from the synthetic likelihood and the prior. We will describe these steps in the following.

\subsection{Choice of emulator}
\label{sec:emulator_choice}

We use neural network based emulators $q(\data|\param; \bphi)$: parameters $\param$ are given as inputs to the network, and the network is trained to approximate $p(\data|\param)$.
In contrast to traditional synthetic likelihood approaches \cite[]{Wood2010}, we are not restricted to using a (multivariate) normal distribution to approximate the conditional density $p(\data|\param)$. The output form of the emulator is chosen according to our knowledge regarding the conditional density of the simulator. In our second example application, we e.g. model $\data|\param$ as a binomial distribution over 8-bit integer pixel values, and in the third example we model a categorical distribution. If the noise model of the simulation process is unknown, flexible conditional density estimators such as conditional autoregressive models~\cite[]{oord2016pixel, papamakarios_masked_2017} can be readily used in our approach.

\subsection{Inference on the parameters of the emulator}
\label{sec:emulator_uncertainty}

We use probabilistic neural networks, i.e. we represent uncertainty about the parameters $\bphi$ of the emulator $q(\data|\param; \bphi)$. We then use these uncertainties to guide the acquisition of training data for the emulator using active learning (as discussed in the next section). 

In the Bayesian framework, uncertainty is represented through the posterior distribution. Multiple approaches for estimating the posterior distributions over neural network parameters have been proposed, including MCMC methods to draw samples from the full posterior \cite[]{Welling_2011, Chen_2014} and  variational methods, e.g.\ using factorising posteriors \cite[]{Blundell_2015} or normalizing flows~\cite[]{louizos2017multiplicative}.  Finally, deep ensemble approaches \cite[]{Lakshminarayanan16} represent predictive distributions through ensembles of networks. They have the advantage of not requiring the choice of a  functional form of the approximation, and are simple to set up.

Our approach can be applied  with any method that represents uncertainty over network parameters. In our experiments, we use deep ensembles to represent uncertainty about $\bphi$, as we found them to combine simplicity with good empirical performance. Instead of training a single emulator network and inferring its posterior distribution, we train an ensemble of $M$ networks with parameters $\{\bphi_m\}_{m=1}^M$.
From here on, we treat $\bphi_m$ as if they were samples from $p(\bphi|\mcD)$, the posterior over network parameters given data. (In practice, these samples will describe local maxima of the posterior.) The posterior-predictive distribution is approximated by $\bbE_{\bphi|\mcD}\big[q(\data|\param, \bphi)\big] \approx \frac{1}{M}\sum_{m=1}^M q(\data|\param; \bphi_m)$.

Networks are trained supervised with data $\mathcal{D}=\big\{(\param_n, \data_n)\big\}_{n=1}^N$. During training, the parameters of the networks are optimized subject to the loss $-\sum_{m=1}^M\sum_{n=1}^{N} \log q(\data_n|\param_n; \bphi_m)$ w.r.t. $\bphi$ \cite[a proper scoring rule as discussed in][]{Lakshminarayanan16}. Networks in the ensemble are initialized differently, and data points are randomly shuffled during training. 

\subsection{Acquisition rules}
\label{sec:acq_rules}

We use active learning to selectively acquire new samples. We distinguish between two scenarios: In the first, we have particular observed data $\obsdata$ available, and train a \emph{local emulator} which approximates the likelihood near $\obsdata$. This approach requires learning a new emulator for each new observed data $\obsdata$.

We also consider a second scenario, in which we learn a \emph{global emulator} -- which approximates $p(\data|\param)$ globally. Learning a global emulator is more challenging and may potentially require more flexible density estimators. However, once the emulator is learned, we can readily approximate the likelihood for \emph{any} $\obsdata$, therefore amortizing the cost of learning the emulator.

The two scenarios call for different acquisition functions for proposing new samples, which we will discuss next.

\subsubsection{Acquisitions for local emulator learning}
\label{sec:acq_local}

With given $\obsdata$, we want to learn a local emulator that allows us to derive a good approximation to the (unnormalized) posterior $\tp(\param|\obsdata) \propto \bbE_{\phi|\mcD}\big[q(\data=\obsdata|\param; \bphi)\big]p(\param)$.

As we are interested in increasing our certainty about the posterior, we target its variance, $\bbV_{\bphi|\mcD}[\tp(\param|\obsdata, \bphi)]$, where $\bbV_{\bphi|\mcD}$ denotes that we take the variance with respect to the posterior over network weights given data $\mcD$. Thus, we use an acquisition rule which targets the region of maximum variance in the predicted (unnormalized) posterior,
\begin{equation}
\label{maxvar}
\param^* = \argmax_{\param} \bbV_{\bphi|\mcD}[\tp(\param|\obsdata, \bphi)] 
		 = \argmax_{\param} \log p(\param) + \log \sqrt{\bbV_{\bphi|\mcD}[\hat{\mathcal{L}}({\param})]}.
\end{equation}
We approximate $\bbV_{\bphi|\mcD}$ with the sample variance across $\bphi_m$ drawn from the posterior over networks. We refer to this rule as the \emph{MaxVar} rule \cite[]{Jarvenpaa2017}. 
We optimize this acquisition rule by using gradient descent, making use of automatic differentiation to take gradients with respect to $\param$ through the synthetic likelihood specified by the emulator.


\subsubsection{Acquisitions for global emulator learning}
\label{sec:acq_global}

A global emulator may be used to do inference once $\obsdata$ becomes available. Here, the goal for active learning is to bring the emulator $q(\data|\param; \bphi)$ close to the simulator $p(\data|\param)$ for all $\param$s
using as few runs of the simulator as possible. We use a rule based on information theory from the active learning literature \cite[]{Houlsby11, Gal17, Depeweg17}.
We refer to the rule
\begin{equation}
        \param^* = \argmax_{\param} \bbI[\data, \bphi | \param, \mcD] 
                 = \argmax_{\param} \bbH[\data | \param, \mcD] - \bbE_{\bphi|\mcD}\big[ \bbH[\data|\param, \bphi] \big]
\end{equation}
as the maximum mutual information rule (\textit{MaxMI}). See \myappendix{acq_global} for  details.


\subsection{Deriving the posterior distribution from the emulator}

Once we have learned the emulator, we use Hamiltonian Monte Carlo \cite[HMC,][]{Neal_2010} to draw samples from our approximate posterior, using the emulator-based synthetic likelihood. We generate samples of $\param$ drawn from the distribution $\tp(\param|\obsdata) = \bbE_{\bphi|\mcD}\big[q(\obsdata|\param)\big] p(\param)$.
In practice, we sample $\param$ from each ensemble member individually and use the union of all samples as a draw from the approximate posterior. We could also obtain the posterior through variational inference, but here prefer to retain full flexibility in the shape of the inferred posterior.

\section{Results}

We demonstrate likelihood-free inference with emulator networks on three examples: i) we show that emulators are competitive with state-of-the-art on an example with Gaussian observations; ii) we demonstrate the ability of emulators to work with high-dimensional observations while learning to amortize the simulator; iii) we show an application from neuroscience, and infer the posterior over parameters of a biophysical neuron model.

\subsection*{i) Low-dimensional example: Simulator with Gaussian observations}
\label{sec:gaussian}

\begin{figure}[t!]
\includegraphics[width=1.0\textwidth]{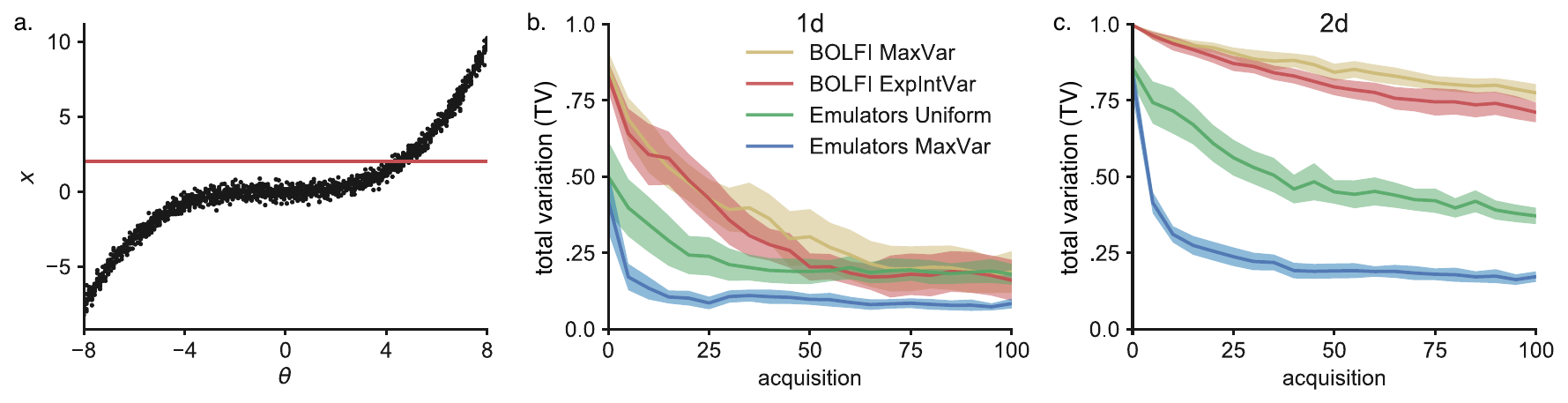}
\vspace*{-10mm}
\caption{
    {\bf Inference on simulator with Gaussian noise}.
    {\bf a.} Data is generated from $\bx \sim \mathcal{N}(\bx|f(\param), \bSigma)$ with cubic non-linearity. We illustrate posterior inference  $p(\param|\obsdata)$ given $\obsdata=2$ (red line at $\obsdata$).
    {\bf b.} In 1-D, emulator-based inference with \textit{MaxVar} acquisitions leads to faster convergence to true posterior than uniform sampling, or BOLFI. Total variation (TV) is measured between true and approximate posteriors. $100$ acquisitions starting from $N_{\text{initial}}=10$ initial points. Lines are means and SEMs from 20 runs. 
    {\bf c.} Same problem, but $\data$ and $\param$ $\in \mathds{R}^2$, non-linearity applied point-wise, starting from $N_{\text{initial}}=25$ points.
\label{fig:gauss} }
\end{figure}

We first demonstrate emulator networks on a non-linear model between parameters and data, corrupted by additive Gaussian observation noise: data is generated  according to $\data_i \sim \mathcal{N}(\cdot|f(\param), \bSigma)$, $i=1 \dots n$, where $f(\param)$ is cubic in $\param$, $\bSigma$ is fixed, and $\param$ is distributed uniformly (see \myappendix{gauss} for complete specification). The goal is to approximate the posterior $p(\param|\bar{\data}_o)$ from a small number of draws from the generative model (\myfig{gauss}{\color{blue}a}). We parameterize $q(\bx|\btheta; \bphi)$ using a Gaussian distribution whose mean and precision are the output of a neural network with one hidden layer consisting of 10 $\mathrm{tanh}$ units.

We will compare our method to BOLFI \cite[Bayesian Optimization for Likelihood-free Inference,][]{Gutmann2018}, an ABC method which -- given a user-specified discrepancy measure -- learns a $\GP$ that models the distribution of discrepancies between summary statistics of $\data$ and $\obsdata$. \cite{Jarvenpaa2017} proposed multiple acquisition rules for BOLFI. The most principled (but also most costly) rule minimizes the expected integrated variance (\textit{ExpIntVar}) of the approximate posterior after acquiring new data. BOLFI is a  state-of-the-art method for simulation-efficient likelihood-free inference, and substantially more efficient than classical rejection-based methods such as rejection-ABC \cite[]{Pritchard1999}, MCMC-ABC \cite[]{Marjoram2003}, SMC-ABC \cite[]{Sisson2007}.  

We use the total variation (TV) between true and approximate posterior (evaluated using numerical integration) to quantify performance as a function of the number of acquisitions. The emulator is trained on an initial dataset and updated after each new acquisition. We find that emulators with \textit{MaxVar} sampling work better than uniform sampling (\myfig{gauss}{\color{blue}b}). Both BOLFI rules (\textit{ExpIntVar} and \textit{MaxVar}) exhibit very similar performance, but require higher number of simulations than emulators to reach low TV values. On a 2-dimensional version of the problem, the qualitative ordering is the same, but the differences between methods are greater (\myfig{gauss}{\color{blue}c}). We did additional runs of BOLFI \textit{MaxVar} to confirm that it eventually converges towards the correct posterior. However, convergence is slow and the quality of the inferred posterior depends strongly on the choice of the threshold parameter used in BOLFI (see \myappendix{bolfi_convergence}).


\subsection*{ii) High-dimensional observations: Inferring the location and contrast of a blob}
\label{sec:binom}

\begin{figure}[t!]
\includegraphics[trim={0 0.7cm 0 0},clip,width=1.0\textwidth]{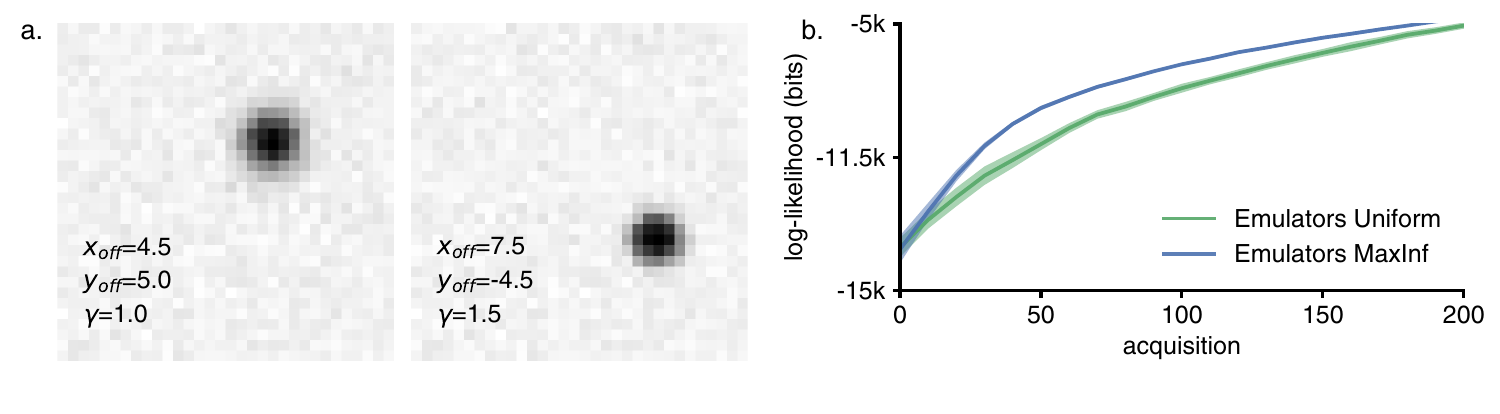}
\vspace*{-6mm}
\caption{
    {\bf Inferring location and contrast of a blob}.
    {\bf a.} Two sample images from the generative model. Parameters are the spatial position and the contrast of the blob.
    {\bf b.} Acquiring samples using the \emph{MaxMI} rule yield to faster emulator learning than samples acquired uniformly in the parameter space. Performances are reported as log-likelihood of held-out test data. $200$ acquisitions starting from $N_{\text{initial}}=50$ initial points. Lines are means and SEMs from 20 runs. 
\label{fig:binom}
}
\end{figure}

We show that our method can be applied to estimation problems with high-dimensional observations without having to resort to using summary statistics. We model the rendering of a blob on a 2D image, and learn a global emulator for the forward model. 

The forward model takes as inputs three parameters ($x_\text{off}$, $y_\text{off}$ and $\gamma$) -- which encode horizontal and vertical displacement, and contrast of the blob -- and returns per-pixels activation probabilities $p_{ij}$. The value of each pixel $v_{ij}$ is then generated according to a binomial distribution with total count $255$ (8-bits gray-scale image) and probability $p_{ij}$, resulting in a $32 \times 32$ pixel image (\myfig{binom}{\color{blue}a}). In this application, we use a multi-layer neural network whose output is, for each pixel, the mean parameter of the binomial distribution (see \myappendix{binom} for further details).

Using the \emph{MaxMI} rule to acquire new test points in parameters space results in faster learning of the emulator, compared to uniform random acquisitions. Eventually, both rules converge towards the log-likelihood of the held-out test set, indicating successful global emulation of the forward model (\myfig{binom}{\color{blue}b}). 
We show posteriors distributions and samples in \myappendix{blob_posteriors}. Since alternative approaches for likelihood-free inference (e.g. BOLFI) do not allow one to globally approximate a simulator, no performance benchmark against these methods was performed.



\subsection*{iii) Scientific application: Hodgkin-Huxley model}
\label{sec:hh}

\begin{figure}[t!]
    \includegraphics[width=1.0\textwidth]{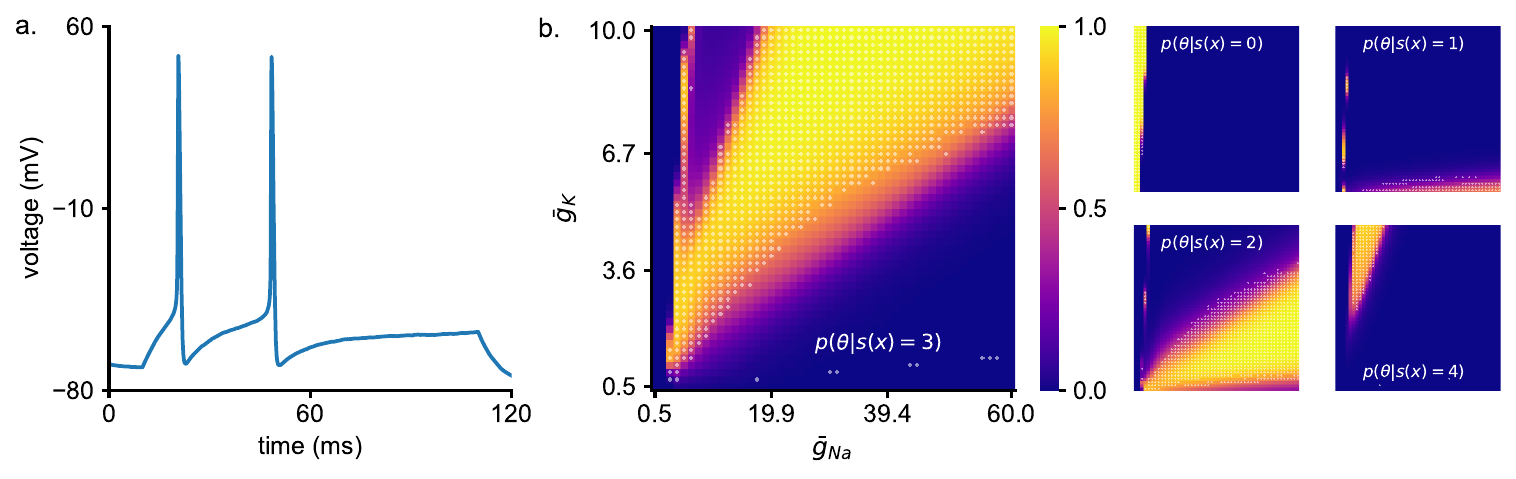}
    \vspace*{-10mm}
    \caption{
        \label{fig:hh}
        {\bf Hodgkin-Huxley model}.
        {\bf a.} Example trace from differential equations describing the model.
        {\bf b.} Posterior inferred for number of spikes as a function of two biophysical parameters. Panels show posteriors for a given number of spikes. The largest panel shows the posterior given three spikes. As a posterior predictive check, we overlay white transparent markers on top of the posteriors where a simulation produced the given number of spikes (and no marker otherwise). 
}
\end{figure}

As an example of a  scientific application, we use the Hodgkin-Huxley model \cite[]{HodgkinHuxley1952} which describes the evolution of membrane potential in neurons (\myfig{hh}{\color{blue}a}). Fitting single- and multi-compartment Hodgkin-Huxley models to neurophysiological data is a central problem in neuroscience, and typically addressed using non-Bayesian approaches based on evolutionary optimization \cite[]{Druckmann2007, VanGeit2016}. In contrast to the previous examples, we do not model the raw data $\data$, but summary features derived from them. While this is often done out of necessity, calculating the posterior relative to summary statistics can be of scientific interest \cite[]{Cornebise2012}. This is indeed the case when fitting biophysical models in neuroscience, which is typically performed with carefully chosen summary statistics representing properties of interest.

Here, we chose to model the number of action potentials (or spikes) in response to a step-current input, and we are interested in the set of parameters that are consistent with the observed number of action potentials. The conditional density of the emulator networks becomes a categorical distribution with 6 classes, modelling the probabilities of exactly 0, 1, \dots 4 spikes, and 5 or more spikes (which never occurred under the parameter ranges we explored). Model parameters $\param$ are the ion-channel conductances $\bar{g}_{\mathrm{Na}}$ and $\bar{g}_{\mathrm{K}}$, controlling the shape and frequency of the spikes (further details in \myappendix{hh}).

We trained emulator networks using $\textit{MaxMI}$ to infer the posterior probabilities over $\param$ generating a given number of observed spikes -- the acquisition surface is shown in \myappendix{hh_acq}. Resulting posterior distributions are shown in \myfig{hh}{\color{blue}b}, along with a posterior predictive check showing that the mapping between parameters and summary features was learned correctly.

\section{Discussion}

We presented an approach for performing statistical inference on simulation-based models which do not permit tractable likelihood. We learn an `emulator network', i.e. a probabilistic model that is consistent with the simulation, and for which likelihoods are tractable. The likelihoods of the emulator can then be plugged into any Bayesian inference approach \cite[as in synthetic likelihood approaches][]{Wood2010, OngNott2016, OngNott2017} to calculate the posterior. Active learning can be used to adaptively suggest new samples to reduce the number of calls to the simulator. We discussed two acquisition functions for learning `local' and `global' emulators, we showed that our approach  scales to high-dimensional observation spaces, does not require user-defined distance functions or acceptance thresholds, and is not limited to Gaussian observations -- all of which are challenging for conventional ABC approaches. 

Our approach uses density estimation to approximate the likelihood. A complementary use of density-estimation in ABC is to directly target the posterior distribution \cite[]{PapamakariosMurray17, Lueckmann2017, LeBaydinZinkovWood2017, Izbicki2018}. This approach can be very useful -- however, one advantage of likelihood-based approaches is that they allow one to apply the same synthetic likelihood to multiple priors (without having to retrain), or to pool information from multiple observations (by multiplying the corresponding synthetic likelihoods). More technically, posterior density estimation gives less flexibility in proposing samples -- in order to yield the correct posterior, samples have to be drawn from the prior, or approaches such as  importance-weighting \cite[]{Lueckmann2017} or other post-hoc corrections \cite[]{PapamakariosMurray17} have to be applied. We discuss additional related work published concurrently with this manuscript in \myappendix{related_work}.

There are multiple ways in which our approach can be improved further: First, one could use alternative, and more expressive neural-network based density estimators, e.g. ones based on normalizing flows \cite[]{papamakarios_masked_2017}. Second, one could use Bayesian posterior estimation (rather than ensembles) to capture parameter uncertainty, and/or use variational inference (rather than HMC) to derive an estimate of the posterior from the synthetic likelihood provided by the emulator. Third, we presented two acquisition functions (one for local and one for global estimation) -- it is likely that the approach can be made more simulation-efficient by using different, and more sophisticated acquisition functions. In particular, our \emph{MaxVar} rule targets the parameters with maximal uncertainty, but does not try to predict whether that uncertainty will be effectively reduced.
However, evaluating acquisition functions like \textit{ExpIntVar} can be computationally expensive -- it will be useful to develop approaches which are sensitive to the relative cost of simulations and proposals, and adaptively adjust the acquisition function used.

Numerical simulations make it possible to model complex phenomena from first principles, and are indispensable tools in many fields in engineering and science. The advent of powerful approaches for statistical inference in simulation-based models \cite[]{BrehmerCranmer2018} is opening up exciting opportunities for closing the gap between mechanistic, simulation-based and statistical approaches to modelling complex systems. Our Bayesian methodology based on emulators provides a fast, effective surrogate model for the intractable likelihood implied by the simulator, and the active-learning based rules lead to bounded-rational decisions about which simulations to run. In combination, they form a rigorous and resource-efficient basis for data analysis with simulators in the loop.


\newpage

\vspace{-0.9cm}
\vspace{-0.9cm}

\newpage

\section*{Acknowledgements}

We thank Marcel Nonnenmacher and Pedro J. Gonçalves for discussions, and help with simulation of the Hodgkin-Huxley model. We thank David Greenberg and all members of the Neural Systems Analysis group for comments on the manuscript.

This work was supported by BMBF (FKZ 01IS18052 A-D, Project ADIMEM) and DFG (SFB 1089, SPP 2041, and SFB 1233, Project 'Robust Vision', 276693517) grants and by the caesar foundation.

\bibliographystyle{plainnat}
\bibliography{refs}
\newpage
\appendix

\section{}
\label{appendix:algorithm}

\begin{algorithm}
  \hrule
  \BlankLine
  \caption{\textbf{ABC via active learning to learn a synthetic likelihood}}
  \label{alg:ours}
  \BlankLine
  \hrule
  \SetAlgoLined
  \SetKwInOut{Input}{Input}
  \SetKwInOut{Output}{Output}
  \DontPrintSemicolon
  \BlankLine
  \Input{  $p(\param)$, $p(\data|\param)$, $\obsdata$ \tcp*{prior, stochastic simulator, observed data} }
  \Output{ $\hat{p}(\param|\obsdata)$ \tcp*{approximate posterior}}
  \BlankLine
  $\mcD \leftarrow {\mcD^{N_\text{initial}} = \big\{(\param_n, \data_n)\big\}^{N_\text{initial}}_{n=1} \sim p(\data, \param)}$ \tcp*{$\param_n \sim p(\param)$, $\data_n \sim p(\data|\param_n)$}
  \BlankLine

  \SetKwRepeat{Do}{do}{while}
  \Do{not converged}{
    Train emulator $q(\data|\param; \bphi)$ on $\mcD$

    Find $\param^{*}$ as the maximum of an acquisition function

    Acquire new data point $\mcD^{*} = \big\{(\param^*, \data^*)\big\}$ by simulating for $\param^*$ \tcp*{$\data^*|\param^* \sim p(\data|\param^*)$}

    $\mcD \leftarrow \mcD \cup \mcD^{\text{*}}$
  }
  \BlankLine
  Find $\hp(\param|\obsdata)$ using the synthetic likelihood $\hlik(\param) = q(\obsdata|\param; \bphi)$
  \BlankLine
  \hrule
\end{algorithm}
\newpage

\section{Acquisition rule for global emulator learning}
\label{appendix:acq_global}

For global emulator learning, we use a rule based on information theory from the active learning literature that maximizes information gain \cite[]{Houlsby11, Gal17, Depeweg17}.
We refer to the rule
\begin{equation}
    \label{eq:MaxMI}
    \begin{aligned}
        \param^* &= \argmax_{\param} \bbI[\data, \bphi | \param, \mcD] \\
                 &= \argmax_{\param} \underbrace{\bbH[\data | \param, \mcD]}_{\text{entropy}}\ - \underbrace{\bbE_{\bphi|\mcD}\big[ \bbH[\data|\param, \bphi] \big]}_{\text{expected conditional entropy}}
    \end{aligned}
\end{equation}
as the maximum mutual information rule (\textit{MaxMI}). 

The first term is the entropy of the data under the posterior-predictive distribution implied by the emulator:
\begin{equation}
    \label{eq:maxent}
    \bbH[\data | \param, \mcD] = - \int \hat{p}(\data | \param, \mcD) \ln \hat{p}(\data | \param, \mcD) \rmd\bx,
\end{equation}
where $\hat{p}(\data | \param, \mcD)$ is obtained by marginalizing out the emulator's parameters w.r.t. $p(\bphi | \mcD)$:
\begin{equation}
    \label{eq:posterior-predictive}
    \hat{p}(\data | \param, \mcD) = \int q(\data|\param,\bphi) p(\bphi|\mcD) \rmd\bphi.
\end{equation}

The expected conditional entropy, $\bbE_{\bphi|\mcD}\big[ \bbH[\data|\param, \bphi] \big]$, is the average entropy of the output $\data$ for a particular choice of inputs $\param$ and emulator parameters $\bphi$, under the posterior distribution of emulator parameters $p(\bphi|\mcD)$. Again, we treat ensemble members $\bphi_m$ as if they were draws from $p(\bphi|\mcD)$.
\citeauthor{Houlsby11} refer to this rule as Bayesian Active Learning by Disagreement (BALD): we query parameters $\param$ where the posterior predictive is very uncertain about the output (entropy is high), but the emulator, conditioned on the value of its parameters $\bphi$, is on average quite certain about the model output (conditional entropy low on average).

For many distributions closed-form expressions of $\bbH\big[\data | \param, \bphi \big]$ are available, but this is in general not true for the entropy of the marginal predictive distribution $\hat{p}(\data | \param, \mcD)$. To overcome this problem, we derived an upper-bound approximation to the entropy term based on the law of total variance: if we characterize the marginal distribution only in terms of its (co)variance $\Sigma_\mcD(\param)$, then $\bbH[\data | \param, \mcD]~\le~\frac{1}{2}\ln\big[(2\pi e)^N |(\Sigma_{\mcD}(\param))| \big]$. Using the law of total (co)variance, we get
\begin{equation}
    \label{eq:total-cov}
    \Sigma_\mcD(\param|\mcD) = \mathrm{Cov}[\data | \param] = \bbE_{\bphi|\mcD}\big[\mathrm{Cov}[\data|\param, \bphi]\big] + \mathrm{Cov}_{\bphi|\mcD}\big[\bbE[\data | \param, \bphi]\big],
\end{equation}
where all expectations can be approximated by samples drawn from $p(\bphi | \mcD)$.

Note that the density of the forward model, $p(\bx|\btheta)$, does not appear in this rule. By using the upper-bound, we can use gradient-based optimization to find $\btheta^*$. Alternatively, entropies could be approximated using sample, which, however, would be slower. 

\bibliographystyle{plainnat}
\bibliography{refs}
\newpage
\section{Additional related work}
\label{appendix:related_work}

\cite{papamakarios_sequential_2018}, concurrently and independently to our approach \cite[][an earlier preprint version of this work]{lueckmann_2018}, proposed learning synthetic likelihoods using neural density estimators for likelihood-free inference: They use Masked Autoregressive Flows as synthetic likelihoods and report state-of-the-art performance compared to methods that directly target the posterior. Like our approach, the density estimator is trained on sequentially chosen simulations. Rather than using acquisition functions that take into account uncertainty to guide sampling, they draw samples from the current estimate of the posterior. Their approach corresponds to an alternative way of learning a local emulator. 

The recent workshop paper of \cite{durkan_2018} compares \cite{papamakarios_sequential_2018} and our approach on three toy problems learning local emulators. On these toy-problems, both methods are similarly efficient (and more efficient than methods directly targeting the posterior), however, the wallclock time of our method is substantially higher, because of the additional cost of evaluating the acquisition function. Whether this additional cost is warranted on a given problem will depend both on any additional gain brought about by the active selection of samples, as well as the cost of the simulator.  For expensive simulation costs, additional computational budget should be spent to carefully decide for which parameters to simulate. 

\bibliographystyle{plainnat}
\bibliography{refs}

\let\cleardoublepage\clearpage

\newpage
\section{Gaussian simulator example}
\label{appendix:gauss}

\subsection{Model}

Data is generated independently according to $\data_i \sim \mathcal{N}(\cdot|f(\param), \bSigma)$, $i=1\dots n$, where $n=10$, $f(\param) = (1.5\ \param+0.5)^3 / 200$, $\bSigma_{ii}=0.1$, $\bSigma_{ij}=0$ for $i \neq j$, $\bar{\bx}_o = \frac{1}{n} \sum_i^n \bar{\bx}^{(i)}_o = \mathbf{2}$, and $\param$ is distributed uniformly in $[-8,8]^p$ where $p$ is the dimensionality of the problem. 

This problem is inspired by the Gaussian example studied in \cite{Jarvenpaa2017}, where $f$ was chosen as  $f(\param) = \param$. We introduce a nonlinearity in $f$, since our method with uniform acquisitions would otherwise trivially generalize across the space -- we observed that a neural network with the right amount of ReLu units can learn the linear mapping perfectly, independently of where the training samples are acquired.

\subsection{Evaluation}

We evaluate our method and BOLFI \cite[]{Jarvenpaa2017} on this problem in $1$D and $2$D. In $1$D, algorithms start with $N_{\text{initial}}=10$ initial samples, in $2$D with $N_{\text{initial}}=25$, and make 100 acquisitions after each of which we evaluate how well the ground truth posterior is recovered. 

As performance metric, we calculate total variation (TV) between $\hp(\param|\obsdata)$ and $p(\param|\obsdata)$, defined as
$$
\frac{1}{2}\ \int \Big| \hp(\param|\obsdata) - p(\param|\obsdata) \Big| \rmd\param.
$$

\subsection{Network architecture and training}

Emulator networks model a normal distribution as output, so that the outputs of the network parametrise mean and covariance (Cholesky factor of the covariance matrix). Neural networks have one hidden layer consisting of 10 $\mathrm{tanh}$ units. We train an ensemble of $M=50$ networks using Adam \cite[]{kingma_2014} with default parameters ($\beta_1=0.9, \beta_2=0.999$) for SGD, and a learning rate of $0.01$. 

\subsection{BOLFI}

BOLFI requires choice of a distance function: We use the the Mahalanobis distance 
$$\Delta_{\param} = \big( (\bar{\data} - \bar{\data}_o)^T \bSigma^{-1} (\bar{\data} - \bar{\data}_o) \big)^{1/2}, $$

in line with the distance function used for the Gaussian example studied in \cite{Jarvenpaa2017}. We use the implementation provided by the authors \cite[]{elfi}. 

\bibliographystyle{plainnat}
\bibliography{refs}

\newpage
\section{Image example}
\label{appendix:binom}

\subsection{Model}

Images are generated according to:
$$
\begin{aligned}
I_{xy} &\sim \text{Bin}(\cdot|255, p_{xy}) \\
p_{xy} &= 0.9 - 0.8 \exp^{-0.5 \big(r_{xy} /\sigma^{2}\big)^\gamma} \\
r_{xy} &= (x - x_{\text{off}})^2 + (y - y_{\text{off}})^2, 
\end{aligned}
$$

where $x$ and $y$ are coordinates in the image, and $\text{Bin}(\cdot|n,p)$ is the binomial distribution.

Model parameters are $x_{\text{off}}$ and $y_{\text{off}}$, which respectively determine the horizontal and the vertical offset of the blob, $\gamma$, defining its contrast, and $\sigma^2$, determining the width.

For our experiments, we use images of size $32\times32$ pixels. We choose uniform priors in the range $[-16, 16]$ for $x_{\text{off}}$ and $y_{\text{off}}$, and a uniform prior in the range $[0.25, 5]$ for $\gamma$. We fix $\sigma$ to 2.

\subsection{Evaluation}

We evaluate different acquisition methods by keeping track of the log-likelihood of a test set consisting of 5000 parameters-image pairs over the course of acquisitions (starting from an initial sample of size $N_{\text{initial}}=50$). 

\subsection{Network architecture and training}

Emulator networks model a binomial distribution as output. Neural networks have two hidden layers (200 units each) with ReLu activation functions. We train an ensemble of $M=25$ networks using Adam \cite[]{kingma_2014} with default parameters ($\beta_1=0.9, \beta_2=0.999$) for SGD with a learning rate of $0.001$. 

\bibliographystyle{plainnat}
\bibliography{refs}
\newpage
\section{Hodgkin-Huxley example}
\label{appendix:hh}

\subsection{Model}

The dynamic equations describing the evolution of the membrane potential and of the gating variables of the neuron are taken from \cite{Pospischil_2008}:
\begin{align*}
    \label{eq:HH}
    C_m \dot{V} &= -(I_\mathrm{leak} + I_\mathrm{Na} + I_\mathrm{K} +I_\mathrm{M} + I_\mathrm{ext})\\
    &= g_{\text{leak}}(E_{\text{leak}} - V) +%
    \bar{g}_{\text{Na}}m^3 h (E_{\text{Na}} - V) +\\%
    &+ \bar g_{\text{K}}n^4(E_\text{K} - V) +%
    \bar g_\text{M}p(E_\text{K} - V) + I_{\text{in}}(t),
\end{align*}
where $C_m$ is membrane capacitance, $V$ the membrane potential, $I_c$ are ionic currents ($c=\{\mathrm{Na, K, M}\}$) and $I_{\text{in}}(t)$ is an externally applied current which we can imagine as the sum of a static bias $I_{\mathrm{bias}}$ and a time-varying zero-mean noise signal $\varepsilon(t)$. $I_{\mathrm{Na}}$ and $I_{\mathrm{K}}$ shape the up- and down-stroke phases of the action potential (spike), $I_{\mathrm{M}}$ is responsible for spike-frequency adaptation, and $I_{\mathrm{leak}}$ is a leak current describing the passive properties of the cell membrane. Each current is in turn expressed as the product of a maximum conductance ($\bar{g}_c$) and the voltage difference between the membrane potential and the reversal potential for that current($E_c$), possibly modulated by zero or more `gating' variables ($m$, $h$, $n$, $p$). 

Each $x \in \{m, h, n, p\}$ evolves according to first order kinetics in the form:
\begin{equation*}
    \dot{x} = \frac{1}{\tau_x(V)}\big(x_{\infty}(V) - x\big)
\end{equation*}

We provide a step current as input.

In our example application, free model parameters are $\bar g_{\text{Na}}$ and $\bar g_{\text{K}}$. We model uniform priors over these parameters: $\bar g_{\text{Na}}$ is between $0.5$ and $60$ and $\bar g_{\text{K}}$ is between $0.5$ and $10$.

\subsection{Evaluation}

We evaluate the posterior obtained through the emulator after $t=250$ acquisitions, starting from an initial sample size $N_{\text{initial}}=30$. As posterior predictive check, we span a grid over the parameter space and compare simulator outputs to the posterior.

\subsection{Network architecture and training}

Emulator networks model a categorical distribution with $K=6$ classes as output. Neural networks have two hidden layer (200 units each) with a ReLu activation functions. We train an ensemble of $M=25$ networks using Adam \cite[]{kingma_2014} with default parameters ($\beta_1=0.9, \beta_2=0.999$) for SGD with a learning rate of $0.001$. 

\bibliographystyle{plainnat}
\bibliography{refs}

\newpage
\section{BOLFI convergence}
\label{appendix:bolfi_convergence}

\begin{figure}[h!]
\centering
\includegraphics[width=0.5\textwidth]{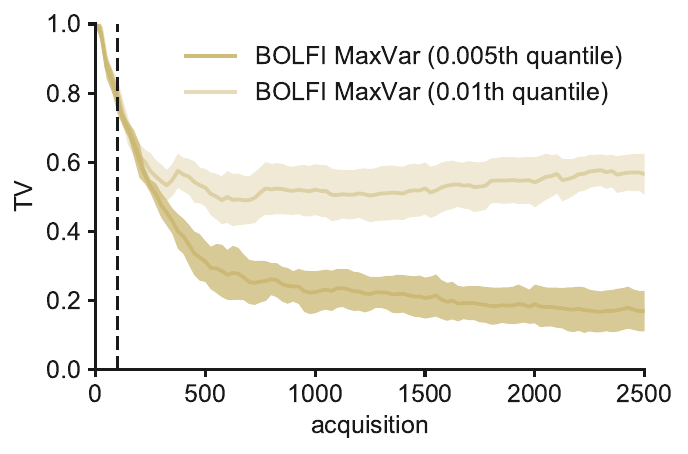}
\caption{
    {\bf Convergence of BOLFI \textit{MaxVar}.} In the manuscript, we show performance up to 100 acquisitions (indicated by the dotted line). With additional acquisitions, BOLFI converges. The quality of the inferred posterior strongly depends on the value of the threshold hyperparameter used in BOLFI.
}
\label{fig:bolfi_convergence}
\end{figure}
\newpage
\section{Posteriors and samples for image example}
\label{appendix:blob_posteriors}

\begin{figure}[h!]
\includegraphics[width=1.0\textwidth]{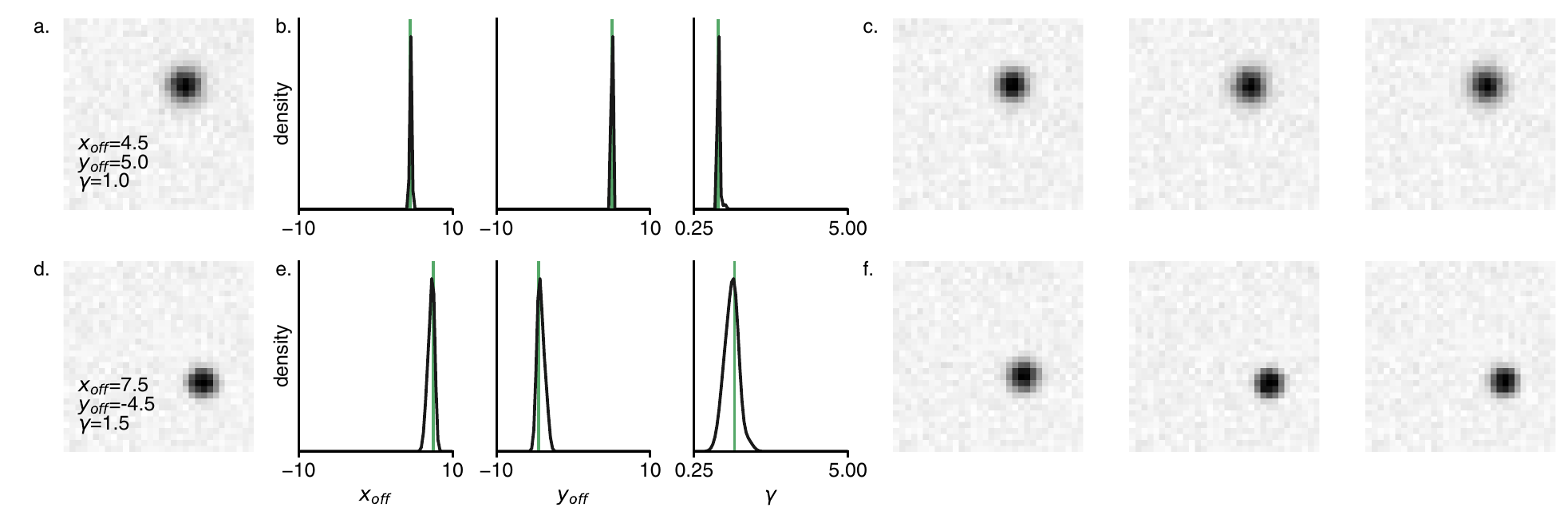}
\caption{
    {\bf Posteriors and samples for image example.} {\bf a.} Observed image. {\bf b.} Inferred posteriors. {\bf c.} Posterior samples. {\bf d.} Another observation, with posteriors in {\bf e.} and samples in {\bf f.}
}
\label{fig:blob_posteriors}
\end{figure}

\newpage
\section{\textit{MaxMI} acquisition for Hodgkin-Huxley model}
\label{appendix:hh_acq}

\begin{figure}[h!]
\includegraphics[width=1.0\textwidth]{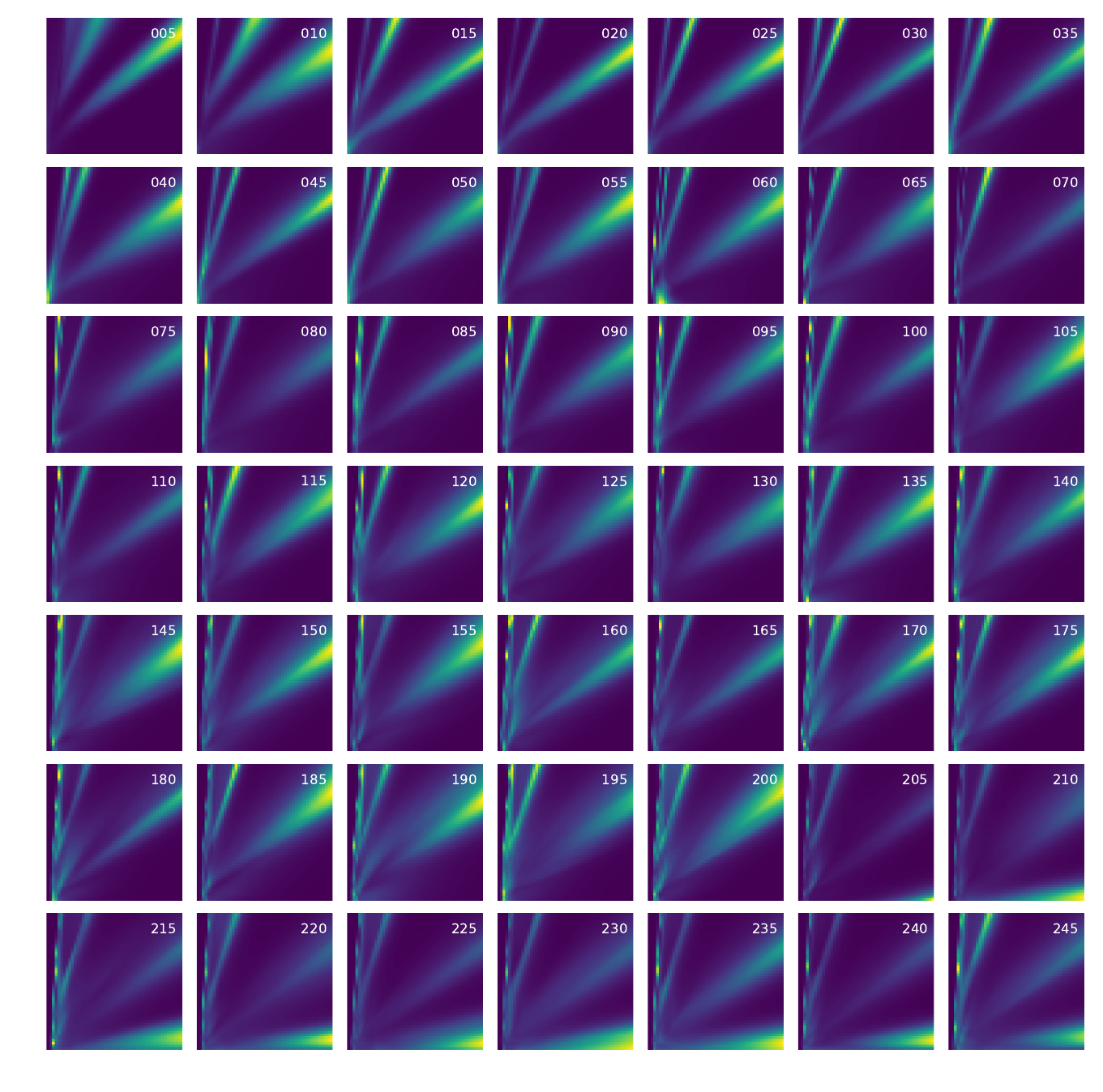}
\caption{
    {\bf Acquistion surface for \textit{MaxMI} rule on Hodgkin-Huxley example.} Individual panels show the acquisition surface over $\param$ as additional samples have been acquired. The acquisition rule proposes datapoints at the decision boundaries of the posterior.  
}
\label{fig:hh_acq}
\end{figure}

\end{document}